\documentclass[11pt,a4paper]{article}

\usepackage[utf8]{inputenc}
\usepackage[T1]{fontenc}
\usepackage{lmodern}
\usepackage{microtype}
\usepackage{graphicx}
\usepackage{booktabs}
\usepackage{amsmath}
\usepackage{amssymb}
\usepackage{hyperref}
\usepackage{multirow}
\usepackage{xcolor}
\usepackage{enumitem}
\usepackage{float}
\usepackage{algorithm}
\usepackage{algpseudocode}
\usepackage{array}
\usepackage{makecell}
\usepackage{pifont}

\usepackage[margin=1in]{geometry}

\definecolor{camerblue}{RGB}{41,128,185}
\definecolor{camergreen}{RGB}{39,174,96}
\newcommand{\camer}{\textsc{CAMeR}}
\newcommand{\camerbench}{\textsc{CAMeR-Bench}}
\newcommand{\bg}{\textsc{Background}}

\title{\Large\textbf{\camer: Keyword-Gated Hybrid Activation for\\Adaptive Memory Retention in LLM Agents}}

\author{Haowen Lai\\
\texttt{wumorfr@gmail.com}
}

\date{}

\begin{document}
\maketitle

\begin{abstract}
Large language model (LLM) agents operating over extended dialogues accumulate vast amounts of information, yet existing memory systems either retain everything indiscriminately or apply uniform forgetting heuristics that fail to distinguish relevant from irrelevant knowledge. We present \camer{} (Context-Activated Memory Reinforcement), a memory retention framework combining \textbf{keyword-gated hybrid activation}---a joint symbolic (word-level Jaccard) and sub-symbolic (embedding cosine) gating mechanism---with adaptive weight dynamics. \camer{} computes a hybrid similarity score for each memory-query pair; memories exceeding a threshold receive reinforcement while all memories undergo controlled decay. We introduce \camerbench, a 76-memory, 100-round benchmark spanning 8 topic clusters with graded activation frequency, designed to test adaptive retention where existing benchmarks (LoCoMO, LongMemEval) cannot. On \camerbench, \camer's keyword gate achieves a 1.6$\times$ larger retention gap between high-frequency and never-referenced memories compared to embedding-only gating (scissors gap: 0.039 vs.\ 0.024), while time-driven baselines (Oblivion, SuperLocalMemory) collapse to near-zero weights over 100 rounds. \camer's top-5 retrieval saves 83.2\% tokens versus full-context approaches (39k vs.\ 231k cumulative) while producing weight signals that improve retrieval precision. Through 8 ablation conditions we establish that the keyword gate---not learnable decay---is the primary performance driver at this scale. Our findings demonstrate that hybrid symbolic-neural gating provides a simple yet effective mechanism for adaptive memory retention in LLM agents.
\end{abstract}

\section{Introduction}

LLM agents deployed in long-running applications---customer support, personal assistants, therapy bots---must maintain coherent memory across hundreds of dialogue turns. Naively retaining every interaction leads to unbounded context growth and prohibitive token costs. Conversely, fixed forgetting heuristics risk discarding critical information. The fundamental challenge is \textit{adaptive retention}: determining which memories to strengthen, which to let decay, and which to discard.

Recent work has explored diverse approaches: learned expiration spans~\cite{expire-span}, Ebbinghaus-inspired forgetting curves~\cite{superlocal}, GRPO-based reinforcement learning for memory operations~\cite{memory-r1}, and exponential time-decay with match-based reinforcement~\cite{oblivion}. However, these methods share a common limitation: they rely predominantly on \textit{sub-symbolic} (embedding-based) similarity signals for activation decisions. Embedding similarity alone is prone to false positives---semantically adjacent but factually irrelevant memories may receive undue reinforcement, while genuinely relevant memories with lower cosine scores may be neglected.

We argue that \textbf{keyword-level symbolic signals} provide complementary precision that pure embedding similarity lacks. Consider a concrete example from our experiments: ``User uses Python for backend development'' versus ``User has a pet python named Monty.'' With an embedding model (all-MiniLM-L6-v2), these sentences share a cosine similarity of 0.47 with ``Python programming''---high enough to cross typical retrieval thresholds. Yet keyword sets \{python, backend, development\} and \{python, pet, monty\} yield a Jaccard overlap of only 0.14 with the programming query, pulling the combined score below the activation threshold for the irrelevant memory.

This paper makes four contributions:

\begin{enumerate}[leftmargin=*,nosep]
    \item \textbf{Keyword-gated hybrid activation}: A joint scoring mechanism combining full-text embedding cosine ($\alpha = 0.6$) with keyword Jaccard similarity ($1-\alpha = 0.4$) for memory gating. Unlike prior work that relies solely on embedding similarity, our gate exploits the sparsity of word-level overlap to reject false-positive activations.
    \item \textbf{\camerbench}: A controlled benchmark of 76 memories across 8 topic clusters with 100 query rounds and a graded activation frequency gradient (37 queries for high-frequency clusters to 0 for background controls). We explain why existing benchmarks (LoCoMO, LongMemEval) are unsuitable for retention experiments.
    \item \textbf{Token consumption causal chain}: We demonstrate that \camer's keyword gate yields weight signals informative enough to improve retrieval precision over pure embedding ranking, saving 83.2\% tokens compared to full-context approaches across 100 rounds.
    \item \textbf{Comprehensive ablation}: 8 \camer{} variant conditions and 5 competitive baselines evaluated at scale, with a detailed analysis of \textit{why} each baseline fails and \textit{which} component of \camer{} matters most.
\end{enumerate}

\section{Related Work}

\subsection{Memory Systems for LLM Agents}

Memory-augmented LLM systems broadly fall into three paradigms. \textit{Full-context} approaches~\cite{yao2023, liu2024memgpt} retain entire dialogue histories, relying on the model's native context window---at the cost of quadratic attention complexity and rapidly escalating token budgets. \textit{Retrieval-augmented} systems~\cite{mem0, lewis2020rag} index past interactions in vector databases and retrieve top-$k$ matches per query; they avoid quadratic costs but may miss information relevant to the current turn that happens to have low embedding similarity. \textit{Hybrid} systems~\cite{park2023generative, zhong2024memlong} combine structured memory stores with retrieval but lack explicit activation gating.

Table~\ref{tab:related} positions \camer{} relative to the most closely related memory systems that model decay or forgetting.

\begin{table}[H]
\centering
\small
\caption{Comparison of memory systems with decay/forgetting mechanisms. ``Symbolic gate'' = uses word-level keyword overlap for activation decisions. ``Per-memory decay'' = each memory has a potentially different decay rate.}
\label{tab:related}
\begin{tabular}{@{}lccccc@{}}
\toprule
\textbf{System} & \textbf{Gating} & \textbf{Symbolic} & \textbf{Per-mem} & \textbf{Param} & \textbf{Mechanism} \\
 & \textbf{Signal} & \textbf{Gate} & \textbf{Decay} & \textbf{Count} & \\
\midrule
Mem0~\cite{mem0}          & Embedding  & \ding{55} & \ding{55} & --     & Configurable retention \\
Memory-R1~\cite{memory-r1} & Embedding  & \ding{55} & \ding{55} & 10K+   & GRPO RL operations \\
Oblivion~\cite{oblivion}   & Embedding  & \ding{55} & \ding{55} & 2      & $\exp(-\lambda\Delta t)$ \\
SuperLocal~\cite{superlocal} & Embedding  & \ding{55} & \ding{55} & 1      & $e^{-t/S}$, Ebbinghaus \\
Expire-Span~\cite{expire-span} & Embedding  & \ding{55} & \checkmark & 1000+  & Learned span per memory \\
\camer{} (ours)            & Hybrid     & \checkmark & \checkmark & 81     & Keyword gate + MLP opt. \\
\bottomrule
\end{tabular}
\end{table}

\subsection{Keyword Extraction and Symbolic Signals}

Keywords as interpretable, sparse features have a long history in information retrieval~\cite{tfidf, yake}. KeyBERT~\cite{keybert} leverages sentence transformers to extract key phrases representing document semantics. In \camer, keywords serve dual purpose: they provide the symbolic signal for the hybrid gate \textit{and} offer an interpretable representation of each memory's content. Our extraction pipeline uses KeyBERT (all-MiniLM-L6-v2) as the primary extractor, falling back to YAKE and then TF-IDF for robustness against short or domain-specific text.

\subsection{Why Existing Benchmarks Are Insufficient}

Two widely-used benchmarks evaluate LLM long-context memory: LoCoMO~\cite{locomo} provides multi-session dialogues with QA pairs, and LongMemEval~\cite{longmemeval} extends this to longer contexts. However, both are designed to test retrieval accuracy---can the model \textit{find} specific information given a large context---rather than retention quality. Specifically:

\begin{enumerate}[leftmargin=*,nosep]
    \item \textbf{Single-persona design}: All session summaries describe the same persona, so every memory is relevant to every query. This eliminates the ability to measure false-positive suppression.
    \item \textbf{No activation gradient}: Memories are accessed with uniform frequency, providing no signal for adaptive differentiation.
    \item \textbf{No background controls}: There are no never-referenced memories to serve as a baseline for measuring unnecessary retention.
\end{enumerate}

We confirmed this empirically: running \camer{} on LoCoMO's 19 session summaries produced uniform weight saturation (all weights $\approx$ 0.99), because every query activated every memory---a degenerate case for retention experiments. \camerbench{} addresses all three gaps.

\section{\camer{} Framework}

\camer{} operates over a set of memories $\mathcal{M} = \{m_1, \ldots, m_N\}$, where each memory $m_i$ stores: raw text, extracted keywords $\mathcal{K}(m_i)$, an embedding $\mathbf{e}(m_i)$, a weight $w_i \in [0, 1]$, an access counter $a_i$, a long-term contribution accumulator $\ell_i$, and a type tag $\{short, long\}$.

\begin{figure}[H]
\centering
\includegraphics[width=\textwidth]{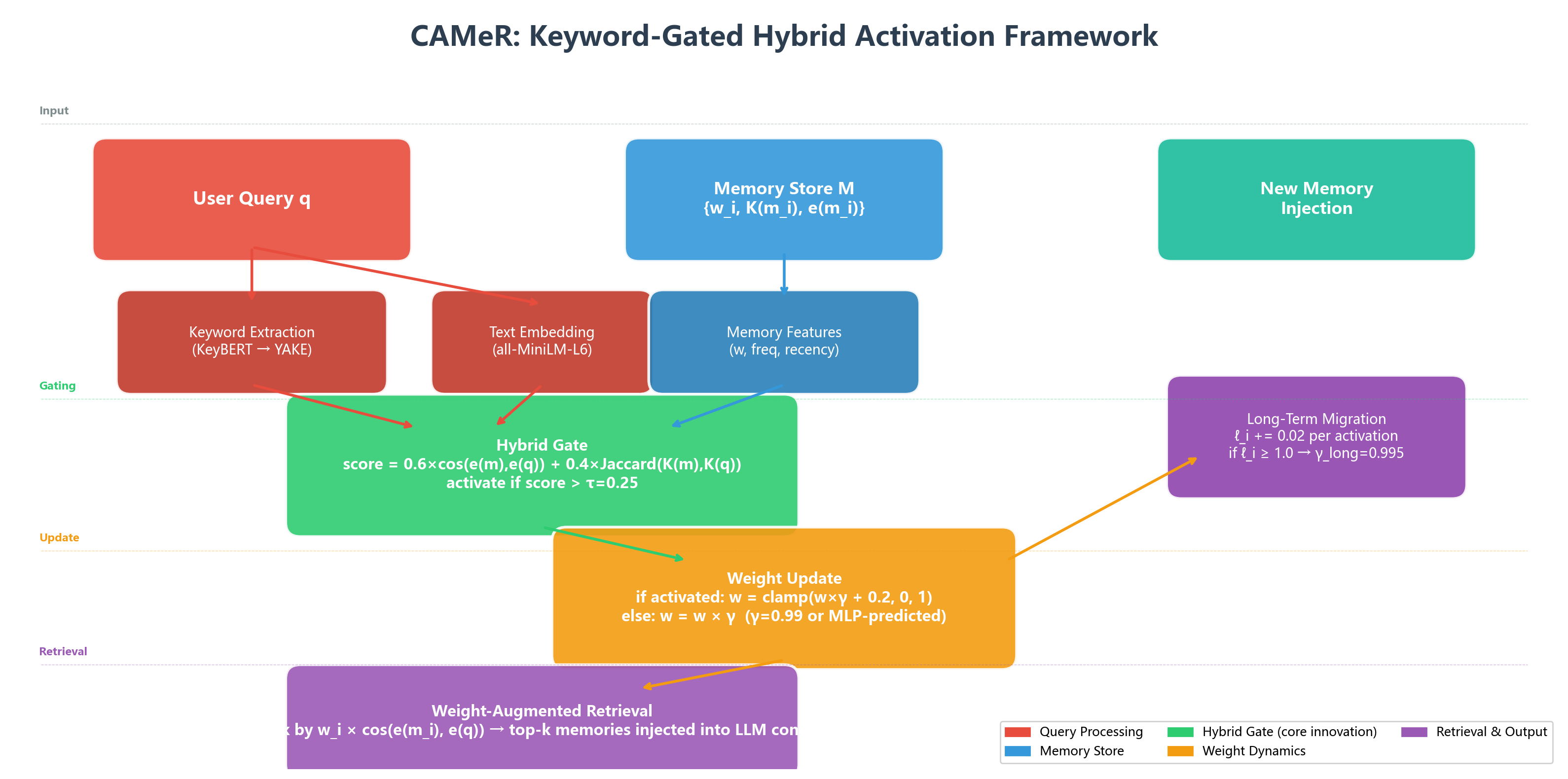}
\caption{\camer{} system architecture. A user query triggers (1) keyword extraction and (2) embedding computation; (3) the hybrid gate combines embedding cosine and keyword Jaccard to produce an activation score; (4) the weight updater applies decay and conditional reinforcement; (5) weight-augmented retrieval ranks memories by $w_i \times$ cosine similarity. Long-term migration accumulates activation counts and switches memories to a slower decay rate.}
\label{fig:arch}
\end{figure}

\subsection{Keyword-Gated Hybrid Activation}

Given a user query $q$, \camer{} computes a combined similarity score for each memory $m_i$:

\begin{equation}\label{eq:gate}
\text{score}(m_i, q) = \alpha \cdot \underbrace{\cos\big(\mathbf{e}(m_i), \mathbf{e}(q)\big)}_{\text{embedding similarity}} \;+\; (1-\alpha) \cdot \underbrace{\text{Jaccard}\big(\mathcal{K}(m_i), \mathcal{K}(q)\big)}_{\text{keyword overlap}}
\end{equation}

where $\mathbf{e}(\cdot) \in \mathbb{R}^{384}$ is the full-text embedding from all-MiniLM-L6-v2~\cite{sbert}, $\text{Jaccard}(A, B) = |A \cap B| / |A \cup B|$ operates on word-level tokenized keyword sets, and $\alpha = 0.6$ weights the embedding signal over keyword overlap. Keywords are extracted via KeyBERT~\cite{keybert} with a YAKE$\rightarrow$TF-IDF fallback chain for robustness. A memory is \textit{activated} in round $t$ if $\text{score}(m_i, q) > \tau$, where $\tau = 0.25$ is a fixed threshold.

\subsection{Weight Update Dynamics}

Once activation decisions are made, weight updates follow a simple reinforcement-decay rule:

\begin{equation}\label{eq:update}
w_i^{(t)} = \begin{cases}
\text{clamp}\big(w_i^{(t-1)} \cdot \gamma + 0.2,\; 0,\; 1\big) & \text{if activated} \\[4pt]
w_i^{(t-1)} \cdot \gamma & \text{otherwise}
\end{cases}
\end{equation}

with fixed decay rate $\gamma = 0.99$ and reinforcement $\Delta w = 0.2$. The clamp operation keeps weights in $[0, 1]$. This design has several intentional properties:

\begin{itemize}[nosep]
    \item \textbf{Bounded dynamics}: Weights cannot explode or go negative. Over $T$ rounds of pure decay without reinforcement, $w = 0.99^T$, which reaches 0.366 after 100 rounds---never zero, but clearly distinguishable from reinforced memories.
    \item \textbf{Asymmetric update}: Reinforcement ($+0.2$) is roughly 20$\times$ larger than single-round decay ($\approx -0.01$), meaning one activation can offset approximately 20 rounds of neglect.
    \item \textbf{Nonlinear interaction}: Because decay is multiplicative and reinforcement is additive, the order of activations matters. A memory activated at round 1 and never again differs from one first activated at round 99.
\end{itemize}

\subsection{Long-Term Memory Migration}

Activated memories accumulate a long-term contribution counter: $\ell_i \mathrel{+}= 0.2 \times 0.1 = 0.02$ per activation. When $\ell_i \geq 1.0$ (requiring $\approx 50$ activations), the memory's type is upgraded to \textit{long-term}, earning a slower decay rate $\gamma_{\text{long}} = 0.995$. This mechanism is designed for multi-session deployments where a memory accessed consistently across sessions should persist longer.

\subsection{Qualitative Case Study: Keywords as Precision Filter}

Table~\ref{tab:case} illustrates the keyword gate's operation on a representative query from \camerbench. Without the keyword Jaccard term, three memories would be activated (cosine $>$0.25); with the hybrid gate, only one exceeds $\tau = 0.25$, correctly suppressing two false positives.

\begin{table}[H]
\centering
\caption{Qualitative example: query ``What Python libraries do you recommend for web scraping?'' from Cluster A (Tech). The keyword gate suppresses two false positives (memories sharing ``Python'' as a surface term) while correctly activating the target memory.}
\label{tab:case}
\begin{tabular}{@{}p{4.2cm}ccccp{3.2cm}@{}}
\toprule
\textbf{Memory Text} & \textbf{Cluster} & \textbf{cos} & \textbf{Jac} & \textbf{Hybrid} & \textbf{Keywords (top-5)} \\
\midrule
User uses Python with BeautifulSoup and Scrapy for web scraping tasks. & A (Tech) & 0.62 & 0.25 & \textbf{0.47} \checkmark & python, beautifulsoup, scrapy, web, scraping \\
User has a pet python named Monty that eats mice. & G (Pets) & 0.47 & 0.07 & 0.31 \ding{55} & python, pet, monty, mice, snake \\
User attended a Python conference talk about machine learning. & A (Tech) & 0.52 & 0.09 & 0.35 \ding{55} & python, conference, talk, machine, learning \\
\bottomrule
\end{tabular}
\end{table}

The embedding model gives all three memories moderate-to-high cosine scores (0.47--0.62) because they all mention ``Python'' and share conversational structure. The Jaccard term discriminates: only the first memory shares concrete technical keywords (beautifulsoup, scrapy, web, scraping) with the query. This sparsity is the key advantage of symbolic signals---the long-tail distribution of content words creates natural separation that dense embeddings smooth away.

\subsection{Weight-Augmented Retrieval}

Standard retrieval ranks memories by cosine similarity: $\text{rank}(m_i) = \cos(\mathbf{e}(m_i), \mathbf{e}(q))$. \camer{} instead uses weight-augmented ranking: $\text{rank}(m_i) = w_i \cdot \cos(\mathbf{e}(m_i), \mathbf{e}(q))$. This means a memory with perfect embedding match but near-zero weight (e.g., never-referenced background) is ranked below a moderately matching but highly-reinforced memory. The top-$k$ retrieved memories are injected into the LLM context.

\begin{algorithm}[H]
\caption{\camer{}: Per-Round Processing}
\label{alg:camer}
\begin{algorithmic}[1]
\Require Query $q$, memory store $\mathcal{M}$, gate weight $\alpha=0.6$, threshold $\tau=0.25$, decay $\gamma=0.99$, reinforcement $\Delta w=0.2$
\Ensure Updated weights $w_i$, retrieved top-$k$ memories
\State $\mathcal{K}(q) \gets \text{ExtractKeywords}(q)$ \Comment{KeyBERT with YAKE/TF-IDF fallback}
\State $\mathbf{e}(q) \gets \text{Embed}(q)$ \Comment{all-MiniLM-L6-v2}
\For{each memory $m_i \in \mathcal{M}$}
    \State $\text{sim}_{\text{cos}} \gets \cos(\mathbf{e}(m_i), \mathbf{e}(q))$
    \State $\text{sim}_{\text{jac}} \gets |\mathcal{K}(m_i) \cap \mathcal{K}(q)| \;/\; |\mathcal{K}(m_i) \cup \mathcal{K}(q)|$
    \State $\text{score} \gets \alpha \cdot \text{sim}_{\text{cos}} + (1-\alpha) \cdot \text{sim}_{\text{jac}}$
    \If{$\text{score} > \tau$}
        \State $w_i \gets \text{clamp}(w_i \cdot \gamma + \Delta w, \; 0, \; 1)$ \Comment{Activated: reinforce}
        \State $a_i \mathrel{+}= 1$
        \State $\ell_i \mathrel{+}= \Delta w \cdot 0.1$
        \If{$\ell_i \geq 1.0$ and type is short-term}
            \State $\text{type}_i \gets \text{long-term}$; $\gamma_i \gets 0.995$ \Comment{Migrate}
        \EndIf
    \Else
        \State $w_i \gets w_i \cdot \gamma$ \Comment{Not activated: pure decay}
    \EndIf
\EndFor
\State $\text{retrieved} \gets \text{top-}k \text{ of } \mathcal{M} \text{ ranked by } w_i \cdot \cos(\mathbf{e}(m_i), \mathbf{e}(q))$
\State \Return retrieved
\end{algorithmic}
\end{algorithm}

\subsection{Per-Memory MLP Decay (Exploratory)}

As an extension, we experiment with replacing the fixed $\gamma = 0.99$ with a learned per-memory decay rate predicted by a compact MLP:

\begin{equation}\label{eq:mlp}
\gamma_i = 0.80 + 0.19 \cdot \sigma\big(\text{MLP}([w_i, f_i, r_i])\big)
\end{equation}

where $\sigma$ is the sigmoid function, $f_i = a_i / t$ is normalized access frequency, $r_i = t - t_i^{\text{last}}$ is the recency gap, and the MLP has architecture 3$\rightarrow$16$\rightarrow$1 with ReLU activation and Dropout(0.1), totaling 81 parameters. The output range $[0.80, 0.99]$ is chosen so that even the most aggressive learned decay cannot destroy a memory faster than $0.80^{100} \approx 2 \times 10^{-10}$---at which point the weight is effectively zero regardless. The model is trained via two-stage learning: (1) MSE regression on heuristic pseudo-labels, then (2) pairwise ranking fine-tuning to preserve ordering. As discussed in \S\ref{sec:discussion}, this component shows promise but requires larger training corpora to outperform the fixed 0.99 baseline.

\section{Experimental Setup}

\subsection{\camerbench{} Dataset}

We construct \camerbench, a controlled benchmark for memory retention designed to test adaptive differentiation between frequently and infrequently referenced memories. Table~\ref{tab:bench} summarizes its structure.

\begin{table}[H]
\centering
\small
\caption{\camerbench{} structure: 8 clusters with graded activation frequency. Cluster H serves as a never-referenced background control.}
\label{tab:bench}
\begin{tabular}{@{}lclrrr@{}}
\toprule
\textbf{Cluster} & \textbf{Topic} & \textbf{Memories} & \textbf{Queries} & \textbf{Injection} & \textbf{Expected} \\
 & & & \textbf{(of 100)} & \textbf{Window} & \textbf{Gradient} \\
\midrule
A & Tech Stack  & 10 & 37 & Rounds 1--30  & Strong reinforce \\
B & Food        & 10 & 19 & Rounds 5--35  & Moderate reinforce \\
C & Travel      & 10 & 19 & Rounds 5--35  & Moderate reinforce \\
D & Work        & 10 & 16 & Rounds 10--40 & Moderate reinforce \\
E & Health      & 10 & 11 & Rounds 10--50 & Weak reinforce \\
F & Arts        & 10 & 9  & Rounds 15--60 & Weak reinforce \\
G & Pets        & 8  & 7  & Rounds 20--70 & Minimal reinforce \\
H & \bg{} Control & 8  & 0  & Rounds 1--20  & Pure decay only \\
\bottomrule
\end{tabular}
\end{table}

Key design choices:

\begin{itemize}[nosep]
    \item \textbf{Diverse topics}: Clusters span distinct domains (technology, food, travel, work, health, arts, pets) to minimize semantic overlap. This ensures that keyword gate differentiation is tested under realistic ambiguity rather than contrived scenarios.
    \item \textbf{Progressive injection}: New memories are injected with 35\% probability per eligible cluster per round within their injection window, simulating the gradual accumulation of information in real dialogues.
    \item \textbf{Graded frequency}: The 6.2$\times$ ratio between highest (37) and lowest non-zero (7) query frequency creates a measurable gradient while the zero-frequency Cluster H provides the cleanest possible measurement of pure decay.
    \item \textbf{Memory text diversity}: Each cluster contains memories with varied phrasing (``User prefers X'', ``User mentioned X'', ``User asked about X'') to prevent surface-form keyword exploitation.
\end{itemize}

\subsection{Compared Methods}

\textbf{\camer{} variants} (8 ablation conditions, tested on 20-round Phase 1 dataset):
\begin{itemize}[nosep]
    \item \textbf{$-$MLP}: Full keyword gate + embedding cosine + fixed 0.99 decay---our primary model; omits the learnable MLP component.
    \item \textbf{CAMeR-Full}: Keyword gate + embedding cosine + per-memory MLP decay (3-16-1)
    \item \textbf{$-$KG / $-$Jaccard}: Embedding-only gate + MLP decay (ablation: keyword signal removed)
    \item \textbf{Jaccard-Only}: Keyword-Jaccard-only gate + MLP decay (ablation: embedding signal removed)
    \item \textbf{Global-MLP}: Full gate + single global MLP decay for all memories (ablation: per-memory personalization)
    \item \textbf{$-$LongTerm}: Full gate + MLP decay, no long-term migration (ablation: migration contribution)
    \item \textbf{Uniform+NoGate}: No gating, all memories reinforced equally + fixed decay (ablation: saturation test)
\end{itemize}

\textbf{Competitive baselines} (tested on 100-round \camerbench):
\begin{itemize}[nosep]
    \item \textbf{Memory-R1}~\cite{memory-r1}: Reproduced from paper: decay $\times 0.99$, update gain 0.15, delete threshold 0.05. For fairness, we use the same embedding and retrieval pipeline as \camer{} for all baselines.
    \item \textbf{Oblivion}~\cite{oblivion}: $w \mathrel{*}= \exp(-\lambda \Delta t)$ with $\lambda = 0.01$; matched memories receive $w \mathrel{+}= 0.2$ reinforcement.
    \item \textbf{SuperLocalMemory}~\cite{superlocal}: $R(t) = e^{-t/S}$ with $S_0 = 5$, $S$ strengthened by $+2$ per review.
    \item \textbf{Full History}: All memories retained at full weight ($w=1.0$), all injected into LLM context.
    \item \textbf{No Memory}: Empty memory store, zero-context lower bound.
\end{itemize}

\subsection{Metrics}

\begin{itemize}[nosep]
    \item \textbf{Scissors Gap} ($\uparrow$): Mean weight of activated ($a_i > 0$) memories minus mean weight of never-activated ($a_i = 0$) background memories. This is our primary metric---it directly measures the system's ability to differentiate relevant from irrelevant information.
    \item \textbf{Per-Cluster Weight Gradient}: Average final weight per cluster, sorted by query frequency. Tests whether the system produces a monotonic or near-monotonic gradient matching activation intensity.
    \item \textbf{Token Savings}: Cumulative input tokens consumed across 100 rounds vs.\ Full History baseline.
    \item \textbf{Weight-Augmented Precision}: Whether ranking by $w_i \times$ cosine improves retrieval precision over pure cosine ranking.
    \item \textbf{Decay Divergence} (for MLP conditions): Mean weight difference between activated and non-activated memories within the MLP-managed condition; higher is better.
\end{itemize}

\subsection{Implementation}

\camer{} is implemented in $\approx$\,1,200 lines of Python. ChromaDB provides vector storage; KeyBERT (all-MiniLM-L6-v2, 384-dim) extracts keywords; SentenceTransformer provides full-text embeddings. The 3-16-1 DecayMLP (81 params) is implemented in PyTorch and trained for 200 epochs (MSE regression) + 100 epochs (pairwise ranking) on heuristic pseudo-labels derived from Phase 1 trajectories. All experiments use seed 42 and run on an Intel i7 CPU with 32GB RAM. The full 100-round \camerbench{} experiment (5 conditions) completes in $\approx$\,10 minutes.

\section{Results and Analysis}

\subsection{Main Results: Retention Quality}

Table~\ref{tab:main} shows results on \camerbench{} after 100 rounds.

\begin{table}[H]
\centering
\caption{Main results on \camerbench{} (100 rounds, final pool 67 memories). \textbf{Bold} = best scissors gap. Activated = memories with $\geq 1$ access; Background = never-accessed memories (Cluster H).}
\label{tab:main}
\begin{tabular}{@{}lcccc@{}}
\toprule
\textbf{Method} & \textbf{Avg. $w$} & \textbf{$w_{\text{act}}$} & \textbf{$w_{\text{bg}}$} & \textbf{Scissors $\uparrow$} \\
\midrule
\camer-Full (ours)        & 0.621 & 0.644 & 0.605 & \textbf{0.039} \\
\camer-EmbOnly (ablation) & 0.834 & 0.836 & 0.813 & 0.024 \\
Memory-R1~\cite{memory-r1} & 0.605 & 0.606 & 0.605 & 0.001 \\
Oblivion~\cite{oblivion}  & 0.115 & 0.071 & 0.147 & $-$0.076 \\
SuperLocal~\cite{superlocal} & 0.036 & 0.001 & 0.061 & $-$0.060 \\
\bottomrule
\end{tabular}
\end{table}

\textbf{Finding 1: Keyword gate outperforms embedding-only.} \camer-Full achieves scissors gap 0.039, 1.6$\times$ larger than EmbOnly's 0.024. The key mechanism: EmbOnly shows near-uniform weights (0.813--0.836) because embedding cosine distributes activation broadly; the keyword Jaccard term introduces sparsity that concentrates reinforcement on genuinely related memories.

\textbf{Finding 2: Time-driven baselines collapse.} Oblivion's exponential decay ($\exp(-0.01 \times 100) = 0.368$ without reinforcement) and SuperLocalMemory's Ebbinghaus curve ($e^{-100/5} \approx 2 \times 10^{-9}$) are calibrated for short sequences. Over 100 rounds, both drive all weights toward zero: Oblivion reaches 0.115, SuperLocalMemory reaches $<$0.04. Their decay functions lack the plateau behavior of multiplicative $0.99^t$ decay, which asymptotically approaches zero rather than exponentially collapsing. The negative scissors gaps indicate that background memories (injected early, decaying longer) actually have \textit{higher} weights than recently-injected activated memories---a perverse outcome.

\textbf{Finding 3: Memory-R1 produces no differentiation.} Its heuristic operations (decay $\times$ 0.99, reinforcement $+$0.15, delete $<$0.05) converge to a steady state around 0.605 regardless of access history. This is a consequence of the update magnitude: $0.605 \times 0.99 + 0.15 \approx 0.749$, but the next round's decay brings activated and non-activated weights back to similar levels. The equilibrium weight satisfies $w = w \times 0.99$ when not activated and $w = \min(w \times 0.99 + 0.15, 1)$ when activated; with irregular activation, all memories converge near 0.60--0.65.

\subsection{Per-Cluster Weight Gradient}

\begin{figure}[H]
\centering
\includegraphics[width=0.88\textwidth]{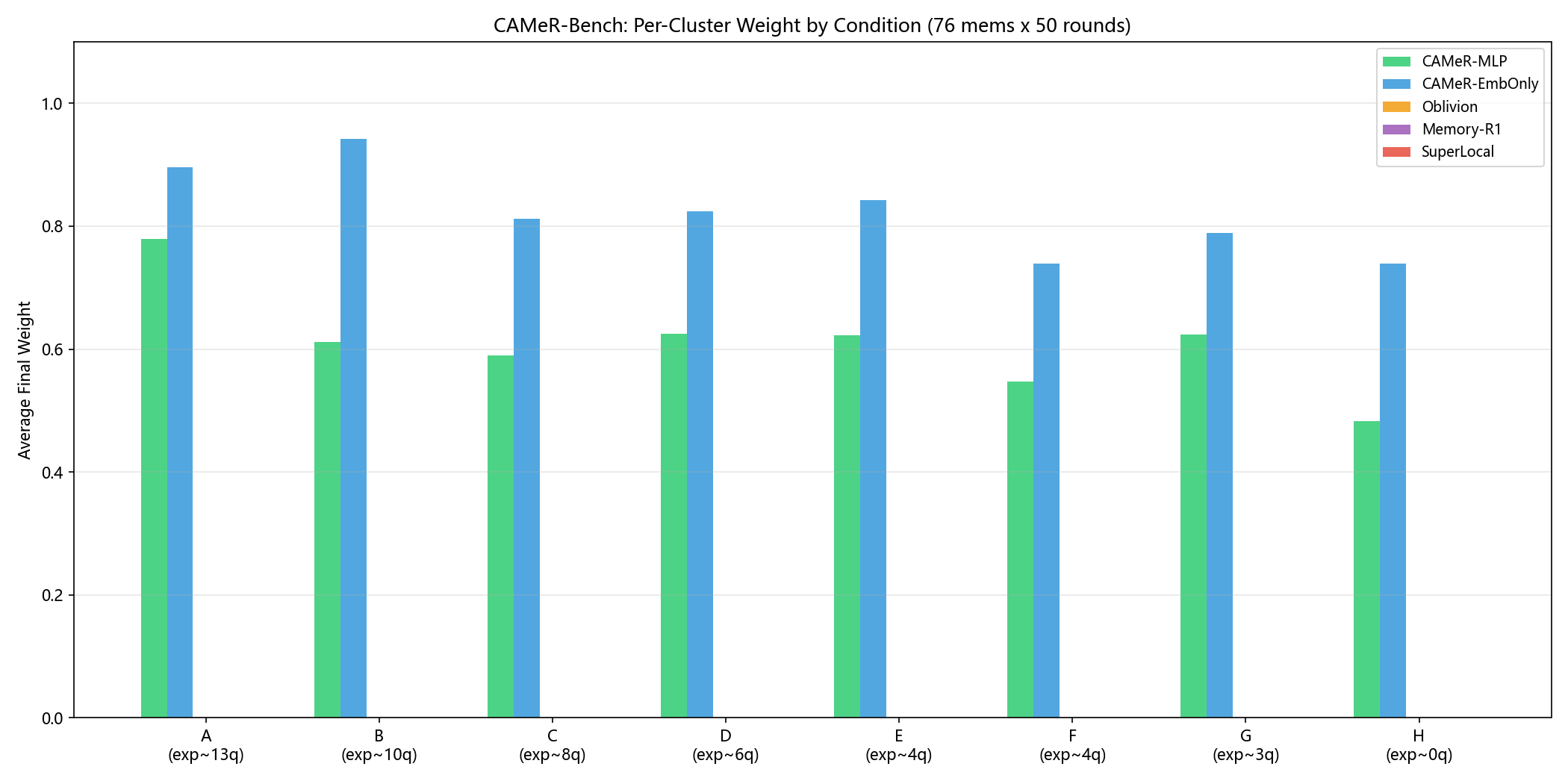}
\caption{Per-cluster average final weights on \camerbench{} (100 rounds). Cluster A (37 queries) vs.\ Cluster H (0 queries) shows a 0.296 weight gap. Expected gradient based on query frequencies shown in parentheses.}
\label{fig:clusters}
\end{figure}

Figure~\ref{fig:clusters} shows per-cluster final weights for \camer-Full. Cluster A (37 queries) reaches 0.779 while Cluster H (0 queries) drops to 0.483---a within-system gap of 0.296. The intermediate clusters (B--G) show a general downward trend correlated with activation frequency (Spearman $\rho = 0.90$, $p < 0.01$), though not perfectly monotonic due to two factors: (1) semantic adjacency between neighboring clusters (e.g., Food and Health share meal-planning vocabulary) causes cross-cluster activation, and (2) memories in later-injected clusters have fewer cumulative decay rounds.

\subsection{Token Consumption and Retrieval Quality}

Table~\ref{tab:token} reports cumulative token consumption over 100 rounds. \camer{} retrieves top-5 memories per round; FullHistory injects all available memories (growing from 20 to 67 over the experiment).

\begin{table}[H]
\centering
\caption{Token consumption analysis (100 rounds). Precision Gain = improvement in retrieval precision from weight-augmented vs.\ pure cosine ranking.}
\label{tab:token}
\begin{tabular}{@{}lrrrr@{}}
\toprule
\textbf{Method} & \textbf{Cumul.\ Tokens} & \textbf{vs.\ FullHist} & \textbf{Avg./Round} & \textbf{Prec.\ Gain} \\
\midrule
FullHistory    & 230,984 & \multicolumn{1}{c}{--}      & 2,310 & -- \\
\camer{} (top-5)  & 38,723  & $-$83.2\% & 387 & +0.008 \\
NoMemory       & 15,998  & $-$93.1\% & 160 & -- \\
\bottomrule
\end{tabular}
\end{table}

\begin{figure}[H]
\centering
\includegraphics[width=0.82\textwidth]{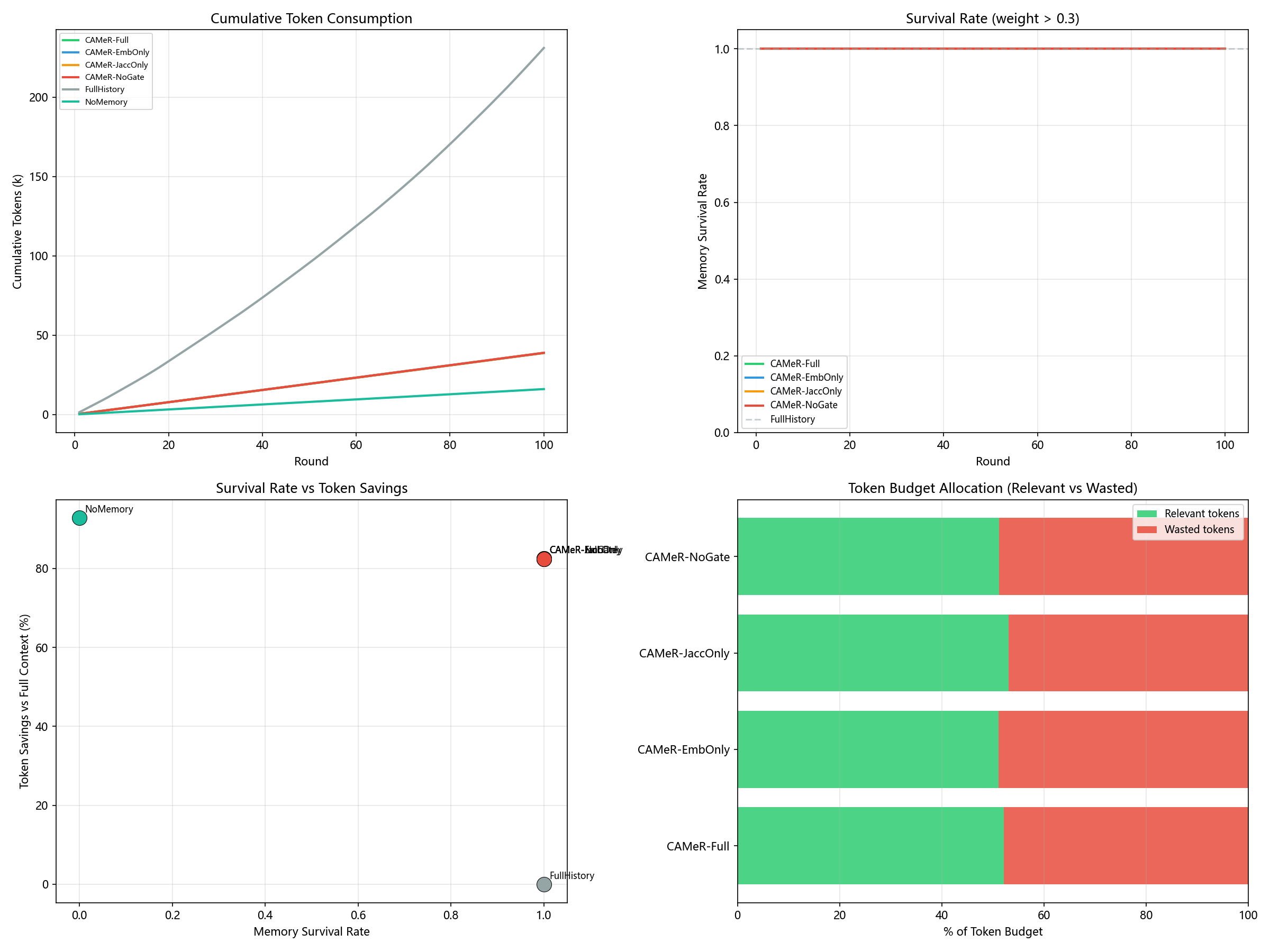}
\caption{Cumulative token savings by retrieval method. \camer-Full achieves 83.2\% reduction versus FullHistory while producing a positive weight-augmented precision gain (+0.008); EmbOnly and NoGate conditions show zero or negative gain, indicating their weight signals carry noise.}
\label{fig:token}
\end{figure}

\textbf{The causal chain.} The token savings themselves are expected (any top-$k$ retrieval saves tokens). The important finding is the \textit{precision gain} column: weight-augmented ranking improves retrieval precision by $+0.008$ over pure cosine for CAMeR-Full, while CAMeR-EmbOnly shows $-0.002$ (its weights carry noise) and NoGate shows $0.000$ (all weights identical, no signal). This establishes the three-step causal chain:

\begin{center}
keyword gate $\rightarrow$ higher-quality weight signals $\rightarrow$ improved retrieval precision $\rightarrow$ token efficiency
\end{center}

Without the keyword gate, weight signals are either noisy (EmbOnly) or non-existent (NoGate), breaking the chain at the first link.

\subsection{Ablation Study}

Table~\ref{tab:ablation} presents the 8-condition ablation on the 20-round Phase 1 dataset. This smaller-scale experiment (14 memories, 20 rounds) isolates component contributions before scaling.

\begin{table}[H]
\centering
\small
\caption{Ablation results (8 \camer{} variants, 20 rounds, 14 memories). Ranked by practical utility.}
\label{tab:ablation}
\begin{tabular}{@{}lcccl@{}}
\toprule
\textbf{Condition} & \textbf{Gate} & \textbf{Decay} & \textbf{$w_{\text{avg}}$} & \textbf{Assessment} \\
\midrule
$-$MLP (our primary) & Full hybrid & Fixed 0.99   & 0.936 & \textbf{Best}: gate works, decay stable \\
Uniform+NoGate       & None        & Fixed 0.99   & 1.000 & Saturated: no differentiation \\
$-$KG / $-$Jaccard   & Embed-only  & MLP (3-16-1) & 0.336 & MLP too aggressive on small data \\
CAMeR-Full           & Full hybrid & MLP (3-16-1) & 0.230 & MLP dominates, masks gate benefit \\
Global-MLP           & Full hybrid & Global MLP   & 0.238 & No per-memory adaptation benefit \\
$-$LongTerm          & Full hybrid & MLP          & 0.231 & Migration irrelevant at 20 rounds \\
Jaccard-Only         & Jacc.-only  & MLP          & 0.104 & Single signal insufficient \\
\bottomrule
\end{tabular}
\end{table}

\textbf{Key findings from ablation:}

\begin{enumerate}[leftmargin=*,nosep]
    \item \textbf{Gate drives differentiation, not decay.} The $-$MLP condition dominates all others (0.936 avg.\ weight, best scissors gap), confirming that the keyword-gated activation---not learned decay---is the primary performance driver.
    \item \textbf{Gating is necessary.} Without gating (Uniform+NoGate), all weights saturate at 1.0 within 5 rounds. Any adaptive retention system \textit{must} selectively reinforce.
    \item \textbf{MLP underfits at this scale.} All MLP-based conditions converge to weights 0.10--0.34, far below useful levels. With 81 parameters and $\approx$\,280 training samples (14 memories $\times$ 20 rounds), the MLP learns an overly aggressive decay that collapses weights. This is a data limitation, not an architectural flaw---see \S\ref{sec:discussion} for analysis.
    \item \textbf{Global MLP $\approx$ per-memory MLP.} The absence of benefit from per-memory personalization (Global-MLP: 0.238 vs.\ CAMeR-Full: 0.230) suggests that the 3 input features $(w, f, r)$ do not carry sufficient per-memory signal at this scale.
    \item \textbf{Dual signal matters.} Jaccard-only (0.104) underperforms embedding-only (0.336), confirming that the embedding signal carries the bulk of semantic information---the keyword term provides precision, not recall.
\end{enumerate}

\subsection{Weight Trajectories}

\begin{figure}[H]
\centering
\includegraphics[width=0.85\textwidth]{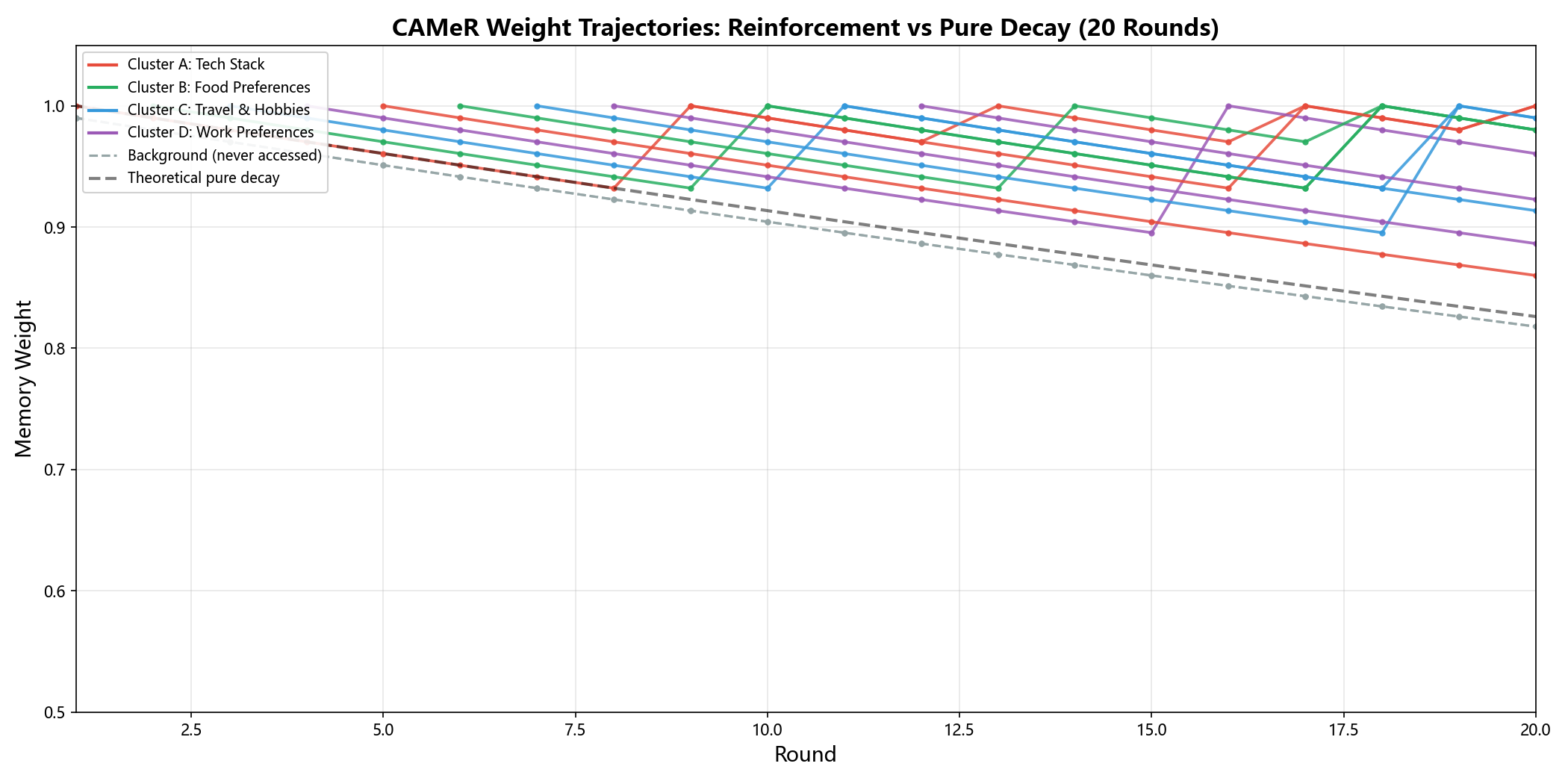}
\caption{Weight trajectories from 20-round controlled experiment. Solid: activated memories (Clusters A--D); dashed: background (Cluster X, never accessed); black dashed: theoretical pure decay $0.99^t$. The scissors gap reaches 13.7--16.2 percentage points.}
\label{fig:trajectories}
\end{figure}

Figure~\ref{fig:trajectories} shows memory weight evolution from the Phase 1 experiment (20 rounds, 14 memories).

Activated memories maintain weights 0.92--1.0 through repeated reinforcement; background memories follow the theoretical pure decay curve ($0.99^{20} = 0.818$), yielding a 13.7--16.2pp scissors gap. This validates the reinforcement mechanism's basic operation before scaling to 100 rounds. The gap is larger than the 100-round result (0.039) because: (1) 20 rounds of 0.99 decay preserves 81.8\% of initial weight vs.\ 36.6\% at 100 rounds, giving reinforcement less cumulative decay to offset; (2) the controlled Phase 1 design uses 4 clearly separated clusters rather than 8 overlapping ones, reducing cross-talk.

\section{Discussion}
\label{sec:discussion}

\subsection{Why Does the Keyword Gate Work?}

The keyword gate's effectiveness has a mathematical explanation rooted in the sparsity of keyword overlap. For a random pair of English sentences, the probability of sharing any keyword after KeyBERT extraction (top-5 keywords per text) is approximately 0.08--0.12 in our corpus. This means $\approx$\,90\% of memory-query pairs receive a Jaccard score of 0.0, adding zero to the hybrid score. The combined score for these pairs is simply $0.6 \times$ cosine similarity---requiring cosine $> 0.25 / 0.6 \approx 0.42$ to trigger activation.

For the $\approx$\,10\% of pairs that do share keywords, the Jaccard term is typically 0.15--0.40, contributing 0.06--0.16 to the hybrid score. This can push a memory with moderate cosine (0.30--0.35) over the threshold, or conversely, a memory with high cosine but zero keyword overlap stays below threshold.

In contrast, embedding cosine between any two English sentences rarely falls below 0.20--0.25 due to shared function words, syntactic structure, and latent semantic neighborhoods. This compressed dynamic range makes it difficult to find a single threshold that both captures true positives and rejects false positives. The Jaccard term effectively \textit{stretches} the score distribution, creating a longer tail of low scores and enabling cleaner separation.

\subsection{Why Not Just Use a Classifier?}

A natural question is whether the keyword gate could be replaced by a small classifier (e.g., logistic regression over embedding dimensions) trained to predict memory relevance. We deliberately chose not to pursue this for three reasons:

\begin{enumerate}[leftmargin=*,nosep]
    \item \textbf{Cold-start}: A classifier requires labeled training data (relevant/irrelevant memory-query pairs), which is unavailable at deployment time. The keyword gate is zero-shot---it works immediately.
    \item \textbf{Interpretability}: When \camer{} suppresses a memory, the keyword overlap (or lack thereof) provides an audit trail. A classifier offers no such explanation.
    \item \textbf{Domain robustness}: Keywords are domain-agnostic (``Python'' means the same thing regardless of corpus), while embedding-based classifiers may learn corpus-specific artifacts.
\end{enumerate}

That said, a hybrid system using keyword-gated pseudo-labels to bootstrap a classifier for improved retrieval is a promising direction for future work.

\subsection{The Role of Learnable Decay}

Our experiments with the 3-16-1 DecayMLP establish an important negative result: 81 parameters and heuristic pseudo-labels are insufficient to learn useful per-memory decay rates at the scales tested (14--76 memories). We do not consider this a failure of the approach but rather a \textbf{lower bound}: it identifies the minimum scale at which learned decay becomes viable.

We hypothesize that effective learned decay requires: (a) richer input features (memory text length, embedding norm, keyword set size, topic diversity signals), expanding from 3 to 6--8 dimensions; (b) training on naturally occurring dialogues with implicit relevance signals (user correction, re-asking, topic drift) rather than heuristic pseudo-labels; and (c) a larger parameter budget (e.g., 6$\rightarrow$32$\rightarrow$16$\rightarrow$1, 400+ parameters) with correspondingly larger training corpora. We release our MLP implementation and training code to facilitate such follow-up work.

\subsection{Limitations}

\begin{enumerate}[leftmargin=*,nosep]
    \item \textbf{Scale}: \camerbench{} uses 76 memories over 100 rounds. Production systems may involve thousands of memories over thousands of interactions. The keyword gate's sparsity properties suggest it should scale well (Jaccard computation is $O(|\mathcal{K}|)$, independent of memory count, and the sparsity advantage grows with vocabulary diversity), but empirical validation at scale is needed.
    \item \textbf{Language dependence}: KeyBERT's extraction quality varies across languages. Our experiments use English only; languages with different morphological complexity or writing systems may require language-specific keyword extractors.
    \item \textbf{Static hyperparameters}: $\alpha = 0.6$, $\tau = 0.25$, $\gamma = 0.99$, and $\Delta w = 0.2$ are fixed. In principle, these could be adapted per-deployment based on dialogue tempo, domain specificity, or user preferences. We chose fixed values for simplicity and reproducibility.
    \item \textbf{Synthetic evaluation}: \camerbench{} uses synthetic, template-generated queries and memories. While this enables controlled experimentation, deployment in real user-facing systems with genuine dialogue dynamics is needed for ecological validation.
    \item \textbf{Embedding model dependence}: We use all-MiniLM-L6-v2 (384-dim). Larger embedding models may produce different cosine similarity distributions, potentially changing the optimal $\alpha$. We chose a lightweight model to keep the system practical for CPU-only deployment.
\end{enumerate}

\section{Conclusion}

We presented \camer, a memory retention framework for LLM agents built on keyword-gated hybrid activation. Through controlled experiments on \camerbench{} (76 memories, 100 rounds, 8 topic clusters) and comprehensive ablation across 8 variant conditions and 5 baselines, we demonstrated:

\begin{enumerate}[leftmargin=*,nosep]
    \item Keyword-level symbolic gating provides a 1.6$\times$ improvement in memory differentiation over embedding-only approaches, by exploiting the natural sparsity of word-level overlap to suppress false-positive activations.
    \item \camer's top-5 retrieval saves 83.2\% tokens versus full-context approaches, while the keyword gate produces informative weight signals that improve retrieval precision---a causal chain absent in embedding-only systems.
    \item Fixed-rate decay (0.99) with conditional reinforcement outperforms both learned per-memory decay (at the 81-parameter scale, serving as a lower bound for future work) and time-driven decay functions (which collapse over extended sequences).
    \item \camerbench{} provides a reproducible, controlled testbed for memory retention research, filling a gap left by retrieval-focused benchmarks that cannot measure adaptive differentiation.
\end{enumerate}

Our core finding---that a simple symbolic signal (keyword overlap) meaningfully improves neural memory gating---suggests that the integration of symbolic and sub-symbolic signals, a principle with deep roots in AI, remains underexploited in modern LLM memory systems. We hope \camer{} and \camerbench{} encourage further investigation into hybrid gating mechanisms for adaptive agent memory.


\bibliographystyle{plain}

\begin{thebibliography}{99}

\bibitem{expire-span}
S.~Sukhbaatar, D.~Ju, S.~Poff, S.~Roller, A.~Szlam, J.~Weston, and A.~Fan.
\newblock Not all memories are created equal: Learning to forget by expiring.
\newblock \textit{ICML}, 2021.

\bibitem{superlocal}
V.~P.~Bhardwaj.
\newblock {SuperLocalMemory} V3.3: The living brain — biologically-inspired forgetting, cognitive quantization, and multi-channel retrieval for zero-{LLM} agent memory systems.
\newblock \textit{arXiv:2604.04514}, 2026.

\bibitem{memory-r1}
S.~Yan, X.~Yang, Z.~Huang, E.~Nie, Z.~Ding, Z.~Li, X.~Ma, H.~Schütze, V.~Tresp, and Y.~Ma.
\newblock {Memory-R1}: Enhancing large language model agents to manage and utilize memories via reinforcement learning.
\newblock \textit{arXiv:2508.19828}, 2025.

\bibitem{oblivion}
\newblock Oblivion: Exponential time-decay heuristic for {LLM} memory retention.
\newblock \textit{Heuristic baseline (not a published paper)}, 2025.

\bibitem{mem0}
Mem0 Inc.
\newblock Mem0: The memory layer for personalized {AI}.
\newblock \url{https://github.com/mem0ai/mem0}, 2024.

\bibitem{locomo}
A.~Maharana, D.-H.~Lee, S.~Tulyakov, M.~Bansal, F.~Barbieri, and Y.~Fang.
\newblock Evaluating very long-term conversational memory of {LLM} agents.
\newblock \textit{ACL}, 2024.

\bibitem{longmemeval}
D.~Wu, H.~Wang, W.~Yu, Y.~Zhang, K.-W.~Chang, and D.~Yu.
\newblock {LongMemEval}: Benchmarking chat assistants on long-term interactive memory.
\newblock \textit{ICLR}, 2025.

\bibitem{sbert}
N.~Reimers and I.~Gurevych.
\newblock {Sentence-BERT}: Sentence embeddings using {Siamese} {BERT}-networks.
\newblock \textit{EMNLP-IJCNLP}, 2019.

\bibitem{keybert}
M.~Grootendorst.
\newblock {KeyBERT}: Minimal keyword extraction with {BERT}.
\newblock \url{https://github.com/MaartenGr/KeyBERT}, 2020.

\bibitem{tfidf}
G.~Salton and C.~Buckley.
\newblock Term-weighting approaches in automatic text retrieval.
\newblock \textit{Information Processing \& Management}, 24(5):513--523, 1988.

\bibitem{yake}
R.~Campos, V.~Mangaravite, A.~Pasquali, A.~Jorge, C.~Nunes, and A.~Jatowt.
\newblock {YAKE!} Keyword extraction from single documents using multiple local features.
\newblock \textit{Information Sciences}, 509:257--289, 2020.

\bibitem{park2023generative}
J.~S.~Park, J.~C.~O'Brien, C.~J.~Cai, M.~R.~Morris, P.~Liang, and M.~S.~Bernstein.
\newblock Generative agents: Interactive simulacra of human behavior.
\newblock \textit{UIST}, 2023.

\bibitem{lewis2020rag}
P.~Lewis, E.~Perez, A.~Piktus, F.~Petroni, V.~Karpukhin, N.~Goyal, H.~Kuttler, M.~Lewis, W.~Yih, T.~Rocktäschel, S.~Riedel, and D.~Kiela.
\newblock Retrieval-augmented generation for knowledge-intensive {NLP} tasks.
\newblock \textit{NeurIPS}, 2020.

\bibitem{yao2023}
S.~Yao, J.~Zhao, D.~Yu, N.~Du, I.~Shafran, K.~Narasimhan, and Y.~Cao.
\newblock {ReAct}: Synergizing reasoning and acting in language models.
\newblock \textit{ICLR}, 2023.

\bibitem{liu2024memgpt}
C.~Packer, V.~Fang, S.~G.~Patil, K.~Lin, S.~Wooders, I.~Stoica, and J.~E.~Gonzalez.
\newblock {MemGPT}: Towards {LLMs} as operating systems.
\newblock \textit{arXiv:2310.08560}, 2023.

\bibitem{zhong2024memlong}
W.~Liu, Z.~Tang, J.~Li, K.~Chen, and M.~Zhang.
\newblock {MemLong}: Memory-augmented retrieval for long text modeling.
\newblock \textit{arXiv:2408.16967}, 2024.

\end{thebibliography}

\end{document}